\documentclass[conference, letterpaper, twoside]{IEEEtran}
\usepackage[letterpaper, left=0.75in, right=0.75in, bottom=0.75in, top=0.75in]{geometry} %margin for conferences
\def\IEEEtitletopspace{18pt} %margin for the first page
\IEEEoverridecommandlockouts

\usepackage[english]{babel}

\usepackage[normalem]{ulem}

\usepackage{amsmath}
\usepackage{layouts}
\usepackage[colorlinks=true, allcolors=blue]{hyperref}
\usepackage{graphicx}
\usepackage{subcaption}
\usepackage{algpseudocode} % for algorithmic environment, looks cooler...
\usepackage{algorithm} % algorithm environment
\usepackage{enumerate} % to allow enumerations with a,b,c in first level
\usepackage{amssymb}
\usepackage{bm} % bold math symbols
\usepackage{cite} % shows multiple citations like [1]-[3]

\usepackage{listings}
\usepackage{array} % smaller text in columns
\newcolumntype{T}{>{\tiny}l} % define a new column type for \tiny
\newcolumntype{H}{>{\Huge}l} % define a new column type for \Huge

\renewcommand{\baselinestretch}{0.915} %{0.9682}

\usepackage[colorinlistoftodos,prependcaption,textsize=tiny]{todonotes}

\makeatletter
\def\ps@IEEEtitlepagestyle{
  \def\@oddhead{\mycopyrightnotice}
  \def\@evenhead{}
}
\def\mycopyrightnotice{
  {\footnotesize
  \begin{minipage}{\textwidth}
  \centering
  \copyright 2022 IEEE. Personal use of this material is permitted. Permission from IEEE must be obtained for all other uses, in any current or future media, including reprinting/republishing this material for advertising or promotional purposes, creating new collective works, for resale or redistribution to servers or lists, or reuse of any copyrighted component of this work in other works. DOI: 10.1109/CASE49997.2022.9926648
  \end{minipage}
  }
}

\newcommand{\titletext}{Capability-based Frameworks for Industrial Robot Skills: a Survey}
\title{\titletext}

\begin{document}
\author{Matteo Pantano, Thomas Eiband, Dongheui Lee %Thomas~Eiband,~\IEEEmembership{Student Member,~IEEE,}
        %and~Dongheui~Lee,~\IEEEmembership{Senior Member,~IEEE}% <-this % stops a space

%\thanks{ %Manuscript received: March xx, 20xx; Revised: May xx, 2022; Accepted: July xx, 2022. This paper was recommended for publication by xxx upon evaluation of the Associate Editor and Reviewers' comments.}
\thanks{M. Pantano is with Siemens Technology, Munich, 81739, Germany \texttt{matteo.pantano@siemens.com}}%
\thanks{T. Eiband and D. Lee are with the Institute of Robotics and Mechatronics, German Aerospace Center (DLR), Wessling, 82234, Germany \texttt{thomas.eiband@dlr.de, dongheui.lee@dlr.de}}% <-this % stops a space
\thanks{M. Pantano, T. Eiband are also with the Chair of Human-centered Assistive Robotics, Technical University of Munich (TUM),  Munich, 80333, Germany}
\thanks{D. Lee is also with Autonomous Systems, Technische Universität Wien (TU Wien), Vienna, 1040, Austria}
\thanks{This work was partially supported by the European Commission’s Horizon 2020 Framework Programme with the project SHOP4CF under grant agreement No 873087. The results obtained in this work reflect only authors view and not the ones of the European Commission, the Commission is not responsible for any use that may be made of the information they contain.}
%\thanks{The work was partially supported by the Helmholtz Association.}%
%\thanks{Digital Object Identifier (DOI): see top of this page.}
}

%\markboth{IEEE ROBOTICS AND AUTOMATION LETTERS, VOL. xx, NO. xx, xxxx 2022}%
%{Pantano \MakeLowercase{\textit{et al.}}: \titletext}
\IEEEoverridecommandlockouts
\IEEEpubid{\makebox[\columnwidth]{978-1-6654-9042-9/22/\$31.00~\copyright2022 IEEE \hfill} \hspace{\columnsep}\makebox[\columnwidth]{ }}
\maketitle
\IEEEpubidadjcol

\begin{abstract}
The research community is puzzled with words like skill, action, atomic unit and others when describing robots' capabilities. However, for giving the possibility to integrate capabilities in industrial scenarios, a standardization of these descriptions is necessary. This work uses a structured review approach to identify commonalities and differences in the research community of robots' skill frameworks. Through this method, 210 papers were analyzed and three main results were obtained. First, the vast majority of authors agree on a taxonomy based on task, skill and primitive. Second, the most investigated robots' capabilities are pick and place. Third, industrial oriented applications focus more on simple robots' capabilities with fixed parameters while ensuring safety aspects. Therefore, this work emphasizes that a taxonomy based on task, skill and primitives should be used by future works to align with existing literature. Moreover, further research is needed in the industrial domain for parametric robots' capabilities while ensuring safety.
\end{abstract}

\begin{IEEEkeywords}
%PLM, PPR, ROS, task, skill, primitive, robot
PPR, HMLV, task, skill, primitive, robot, review, survey
\end{IEEEkeywords}

\section{Introduction}
Detailed a-priori planning of manufacturing processes defined in nowadays industry is going to be soon outdated with "high mix - low volume" (HMLV) manufacturing driven by heterogeneous demand for product variants \cite{Anshuk.2014, Fechter.2018}. Therefore, capability-based engineering envisioned in Industrie 4.0 is slowly entering the domain of manufacturing to ensure business continuity \cite{bayha2020describing}. Through this concept, factories of the future (FoF) will be able to adapt their production plans during order execution as long required capabilities will be used to describe production processes instead of actual resources \cite{bayha2020describing}. However, for the implementation of capability-based production, resources (e.g., robots, CNC machines) will need to provide descriptions of their capabilities (e.g., 3-axis milling, Cold Metal Transfer (CMT) welding) to ensure correct planning by linking their capabilities with manufacturing requirements \cite{bayha2020describing}. One approach to link resources to requirements was initially introduced in the standardized Product Process and Resources (PPR) model described in \cite{CuttingDecelle.2007}. Since then, different Product Lifecycle Management (PLM) systems started using the model for their simulations. However, such definitions have not widely reached the robotics community and different varieties of definitions and nomenclature have been proposed \cite{backhaus2017digital}. Therefore, an alignment between robotic research literature and PPR literature is needed to overcome one of the adoption barriers of capabilities in manufacturing \cite{Froschauer.4262022}. To bridge this gap, this research presents a structured literature review which focuses on definitions used when describing robots' capabilities considering industrial scenarios. More specifically, this work aims to answer the following research questions:
\begin{itemize}
    \item[] \textit{RQ1}: Which nomenclature is most frequent in robotics when describing capabilities? And with which taxonomy?
    \item[] \textit{RQ2}: What are the most investigated robots' capabilities?
    \item[] \textit{RQ3}: What distinguishes industrial robotics applications using robotic capabilities from academic ones?
\end{itemize}  
To provide a clear description of the research done for answering these questions, this work is structured as follows. At first, in Sec. \ref{sec:definitions}, definitions for a capability based architecture are given. Second, in Sec. \ref{sec:reviewprocess}, the structured review criteria are outlined. Third, in Sec. \ref{sec:analysisprocess}, the results from the review are presented. Finally, in Sec. \ref{sec:discussion} and \ref{sec:conclusion}, the conclusions with future outlooks are given. Moreover, all the data used for this review is available in the supplementary on GitHub\footnote{https://github.com/teiband/industrial-skill-review}.

\begin{figure}
\centering
\includegraphics[width=0.9\linewidth]{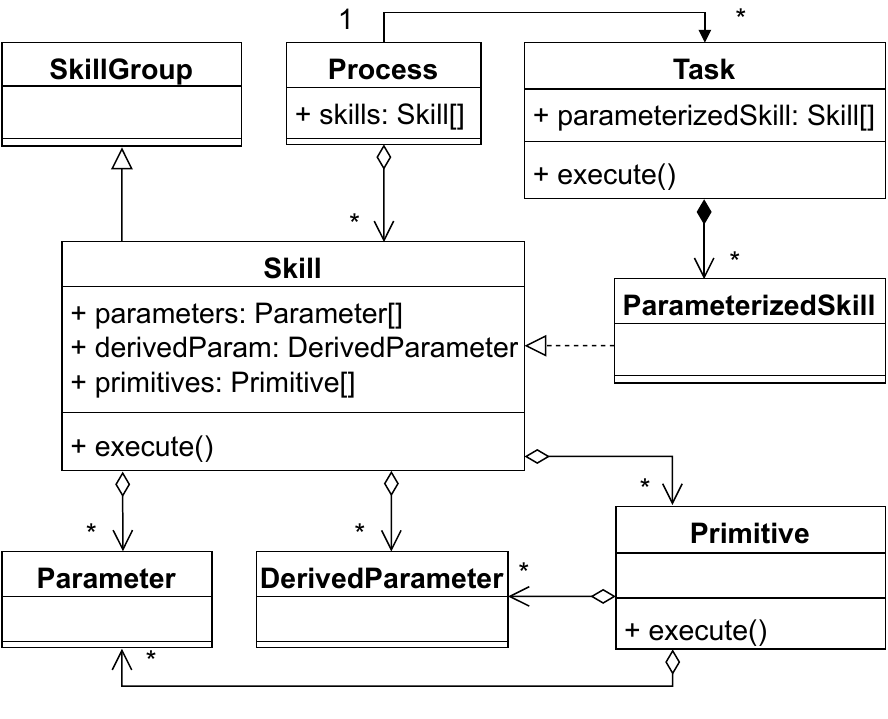}
\caption{Architecture of the capability-based framework used in this review work to conduct the systematic literature review. The figure shows the hierarchy and the relations between the expressions. The expressions were obtained analyzing the literature on PPR and the domain of robotics research.}
\label{fig:skillDiagram}
\end{figure}

\section{Architecture and definitions of terms \label{sec:definitions}}
To conduct the research, the systematic literature review approach defined by \cite{Booth.2021} was employed. Therefore, categories for the classification had to be defined. In this section, such terms are defined. The designated expressions come from two research topics. On one hand, from the manufacturing domain where the concept of Plug-and-Work based on PPR \cite{Schleipen.2015} describes capabilities at the shop floor. On the other hand, from the robotics domain, where ontologies have been defined to represent robots' capabilities necessary to solve complex steps like the assembly of a chainsaw as described in the FoF ontology \cite{schafer2021flexible}. This resulted in the architecture shown in Fig.~\ref{fig:skillDiagram}. 
\begin{figure*}
    \centering
    \includegraphics[width=0.92\textwidth]{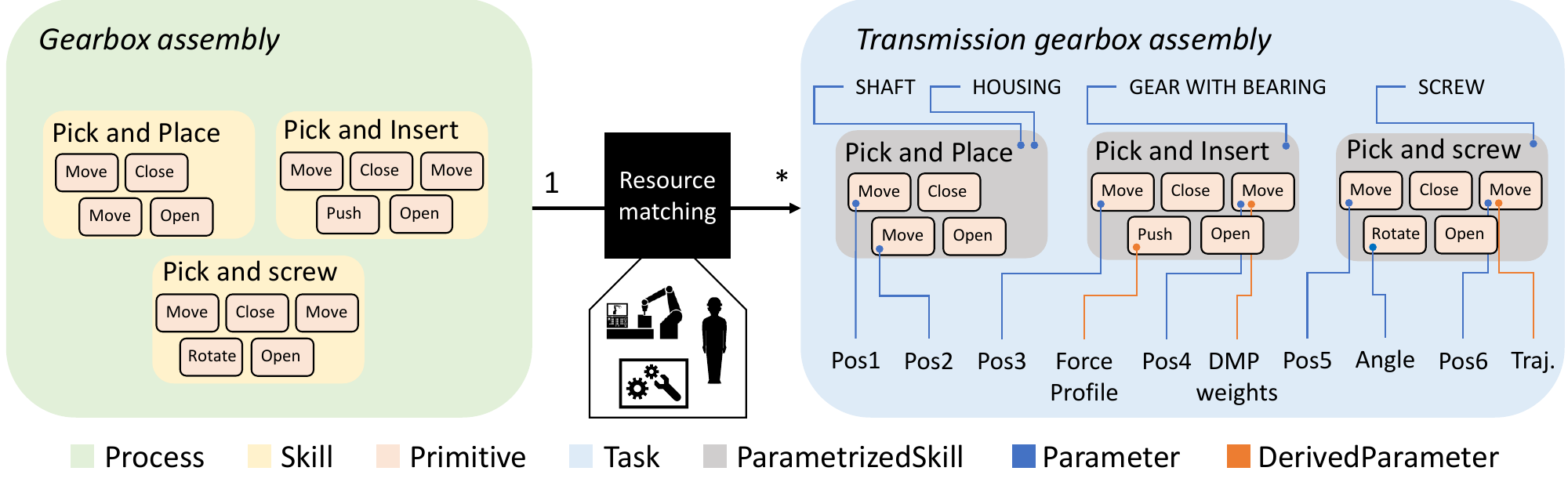}
    \caption{Example of a transmission gearbox assembly. On the left, the general gearbox assembly \texttt{Process} is shown, which is not parameterized and resource independent. The process is composed by the \texttt{Skill}s \emph{Pick and Place}, \emph{Pick and Insert}, and \emph{Pick and Screw}. To the right, a specialized task for a particular transmission gearbox assembly is shown. The task is created by matching resources to a process and specifying parameter values for \texttt{Skill}s and \texttt{Primitive}s. In this case, an object centered representation is used \cite{andersen2014definition, huckaby2012taxonomic, ji2021learning}, the \texttt{Skill} parameter represents the digital artifact of a physical object which contains object specific information (i.e., position). {\scshape Shaft} and {\scshape housing} are passed as parameters to \emph{Pick and Place}. {\scshape Gear with bearing} is passed to \emph{Pick and Insert} and {\scshape Screw} to \emph{Pick and screw}. The artifact's properties can be used to assign parameters also to the underlying \texttt{Primitive}s, for example passing an object's target position to a move. }
    \label{fig:exampleUsageFramework}
\end{figure*}
%\subsection{Definition of Terms}
\subsection{Process}
A \texttt{Process} is defined as an ensemble of different \texttt{Skill}s and depicts an abstract description of steps in a workflow to reach a certain desired outcome as defined by \cite{backhaus2017digital, huckaby2012taxonomic}. Definition of a \texttt{Process} finds its roots in the definitions of the enterprise-control system integration in the well known ANSI/ISA95 \cite{InternationalElectrotechnicalCommission.2013}. However, in this context the definitions of Process as defined in PPR is used. Therefore, a \texttt{Process} is solution neutral and its execution depends on the type of resources involved and their capabilities \cite{backhaus2017digital, Schleipen.2015}.
\subsection{Task}
A \texttt{Task} is defined as an ordered ensemble of different \texttt{Skill}s and depicts a concrete representation of steps in a workflow to solve a specific goal by interfacing with operators, control systems and programs \cite{pedersen2016robot}. Therefore, it could be seen as a more specialized version of a \texttt{Process}. For the sake of clarity, a \texttt{Task} can be a easily described as a sequence of \texttt{Skill}s that have been properly parameterized upon the resources involved and their capabilities. 
\subsection{Skill Group}
A \texttt{SkillGroup} is defined as a collection of \texttt{Skill}s, which allows to group similar ones together. Such grouping has been used in \cite{backhaus2017digital, huckaby2012taxonomic} to structure a large variety of \texttt{Skill}s into meaningful groups that are understandable to the user (e.g. move, connect, compare). The \texttt{SkillGroup} is not considered during execution, but it has a descriptive character when a user is searching for available \texttt{Skill}s. 
\subsection{Skill}
A \texttt{Skill} is a predefined robot's capability that can be parameterized to solve a specific goal. A \texttt{Skill} can be either a physical capability or a perception capability \cite{pedersen2016robot}. \texttt{Skill}s that execute physical actions are able to alter the physical world state, for example picking an object. \texttt{Skill}s with perception capabilities can update only a world representation based on the made observations but do not alter the physical world. An example is the measurement of an object's pose. 
\subsection{Parameterized Skill}
A \texttt{ParameterizedSkill} can be the instance or be implemented as inherited class of a \texttt{Skill} equipped with parameters that are \texttt{Task} and resource specific, hence, it can be executed on robot hardware to accomplish a goal.
\subsection{Parameter}
\texttt{Parameter}s are used to configure a \texttt{Skill} for a specific \texttt{Task} \cite{andersen2014definition,profanter2019hardware}. \texttt{Parameter}s can be specified by different methodologies, for instance manually defined and interpretable by the user, or automatically extracted by the system and non-human interpretable \cite{steinmetz2018razer}. This difference is denoted by calling them respectively \texttt{Parameter} and \texttt{DerivedParameter}.
\subsection{Primitives}
A \texttt{Primitive} is the closest atomic unit to the hardware-level, also know as atomic function that can perform a distinct operation. It can be depicted as building block when composing \texttt{Skill}s, for instance opening a gripper \cite{pedersen2016robot, R.H.Andersen.2014}. Similarly to \texttt{Skill}s, \texttt{Primitive}s can be parameterized to solve a precise task and could provide output information, for example, the location of an object. Additionally, \texttt{Primitive}s can be grouped as \texttt{Skill}s in \texttt{SkillGroup}s, however, this is not considered in this survey.

\subsection{Example of how the architecture can be used}
To demonstrate the proposed architecture an example of an automation process that is solved by a robotic task and how it could be depicted using the above nomenclature is here presented. The example is shown in Fig.~\ref{fig:exampleUsageFramework}. Imagine that the \texttt{Process} of a gearbox assembly shall be automated. Therefore, the user identifies, via a programming method (e.g., learning from demonstration), appropriate robot \texttt{Skill}s (i.e., \emph{Pick and Place}, \emph{Pick and Insert} and \emph{Pick and Screw}) from different \texttt{SkillGroup}s, such as Manipulation. Afterwards, to execute it, resources are matched to the \texttt{Process} via a task planning algorithm or by a capability-based manual assignment on a Manufacturing Execution System (MES) considering that a \emph{Transmission gearbox} needs to be assembled. Therefore, the actual gearbox assembly becomes a \texttt{Task} consisting of a sequence of \texttt{ParameterizedSkill}s. % with assigned parameter values, for example \texttt{Pick and Insert(object=Gear with bearing)}.
These \texttt{ParameterizedSkill}s are the instances of the \texttt{Skill}s equipped with their parameter values. For example, \texttt{Pick and Place(Shaft, Housing)} denotes a \emph{Pick and Place} skill that involves the {\scshape Shaft} and {\scshape Housing} \texttt{Parameters} which are digital artifacts representing properties of the physical objects. The information within these artifacts is then used to assign parameters to the underlying \texttt{Primitives}, for example \texttt{Move(target=pos\_4)}. Furthermore, a \texttt{Move} primitive could also have a \texttt{DerivedParameter} as used within the \emph{Pick and Insert} skill, for example the Dynamic Movement Primitive (DMP) \cite{Schaal.2006} weights of the represented motion, \texttt{Move(target=pos\_4, traj=dmp\_weights)}.

\section{Review process \label{sec:reviewprocess}}
Due to the research bulk on the topic, this review used a systematic research method (SRM) as outlined in Sec. \ref{sec:definitions}. This section highlights how the expression previously defined were used along with exclusion criteria, search strategy and the research protocol.

\subsection{Literature search}
To collect candidate papers, an automatic search on the Scopus digital library database\footnote{https://www.scopus.com/} was performed on 6th October 2021. The search terms targeted the industrial usage of robots' capabilities published between 2014 and today as long this time span presented the largest amount of research publication on the field. This resulted in the following Scopus search string:
\begin{lstlisting}
[(robot AND skill) OR (robot W/15 skill)] AND (industrial OR manufacturing)
\end{lstlisting}
The query was applied to the research fields: title, abstract and keywords and resulted in a total of 210 papers.
\subsection{Selection}
Afterwards, an exclusion criterion on the abstract and title, filtering out papers that were not fitting due to topic irrelevance was applied (i.e., skills in the workforce required for usage of robotics) and 149 papers were discarded. The remaining 61 papers were fully read and analyzed. During the process, 27 other publications were added as long they were describing relevant previous works of authors identified in the previous step. This resulted in a total of 88 fully analyzed papers.
\subsection{Classification}
To understand how and which nomenclature the research papers used to define robotic capabilities, these classification criteria were created:
\begin{itemize}
    \item \textit{Skill model}. Evaluation whether the authors define what a skills is and how a skill model is structured.
    %Defining what a skill is, can be important to understand what the authors intended. If a clear skill model was given by the authors, this term was marked.
    \item \textit{Similarity}. Understanding if the proposed capability-based skill framework is similar to the one presented in order to evaluate the proposal of this work. This criteria recorded if the framework showed the same structure as the one presented in this work.
    \item \textit{Industrial}. To know if the research was more industrially focused or not is important to understand the technical readiness level (TRL) of the technology. Therefore, it was marked if the research work was conducted on a use case of a real-manufacturing scenario.
    \item \textit{Industrial requirements}. Knowing if the requirements for industrial application that are necessary to enter a specific market are met is another important insight to understand the TRL of the technology. Considering that large amount of the literature on capabilities is from Europe \cite{Froschauer.4262022}, the criteria for accessing the European market were used to asses the development level of the frameworks \footnote{similar requirements however, apply also to other markets} (the full list of the requirements can be found in the supplementary material). % The applicability of the technology was surveyed, 
    \item \textit{Implementation}. Knowing the implementation technologies is important to understand the applicability in other scenarios. In this term, the frameworks and programming languages were recorded if implementation details were provided.
    \item \textit{Parameters}. Assigning parameters to skills enables generalization capabilities. If parameters were used, their type was reported.
    \item \textit{Definitions}. To understand how the definitions provided in Sec. \ref{sec:definitions} were used, the nomenclature used in the reviewed works was mapped to the definitions provided above using a review table (the full review table can be found in the supplementary material).
\end{itemize}

\section{Analysis of the results \label{sec:analysisprocess}}
To analyze the results the semantic properties and frequencies of the terms were analyzed. This section describes the results obtained from this analysis.

\subsection{Classifications results}
By applying the classification criteria on the 88 papers the following results were obtained:
\begin{itemize}
    \item \textit{Skill model}. 26 papers out of 88 proposed a clear skill model used in their skill framework.
    \item \textit{Similarity}. 57 papers out of 88 used a capability-based skill framework similar to the one proposed in Sec. \ref{sec:definitions}.
    \item \textit{Industrial}. 45 papers out of 88 were dealing with an industrial use case.
    \item \textit{Industrial requirements}. 61 papers out of 88 considered some of the requirements needed for industrial usage.
    \item \textit{Implementation}. 49 papers out of 88 clearly explained the used tools and frameworks for the implementation.
    \item \textit{Parameters}. 32 papers out of 88 defined and explained the parameters used for their skill frameworks.
    \item \textit{Definitions}. Considering the definitions in Sec. \ref{sec:definitions}, the research papers could be summarized as follows. The categories which had the most amount of information were skill (79 out of 88), task (65 out of 88) and primitives (49 out of 88). The remaining categories were used much less frequently, skill group (14 out of 88), parametrized skill (17 out of 88) and process (31 out of 88).
\end{itemize}

\subsection{Nomenclature}
Within the definitions in Sec. \ref{sec:definitions}, also the types of skills, tasks and primitives were recorded. To study which names were most common across the research works, and provide data for \textit{RQ1} and \textit{RQ2}, the data was preprocessed and the most common terms identified.

\subsubsection{Preprocessing}
In order to prepare the extracted data for clustering, a number of preprocessing steps to the manually extracted definitions obtained from the analysis performed in Sec. \ref{sec:reviewprocess} were applied. For the sake of clarity the denomination of a task, skill, or primitive is defined as label in the following paragraphs. First, the labels from the review table under the definitions column were extracted. Whenever authors provided labels in camel case, they were resolved to words with underscores, for instance \emph{MoveTo} resulted in \emph{Move\_To}. Next, labels were converted to lowercase. Then, lemmatization was applied on each of the words. Here, inflectional endings were removed, i.e., "moving" would result in "move". Hereby, the WordNetLemmatizer from the natural language toolkit (NLTK) \cite{loper2002nltk} was used and 526 labels were obtained for the subsequent steps.

\subsubsection{Identified Common Terms\label{sec:common_terms}}
\begin{figure*}
    \centering
    \includegraphics[width=0.92\textwidth]{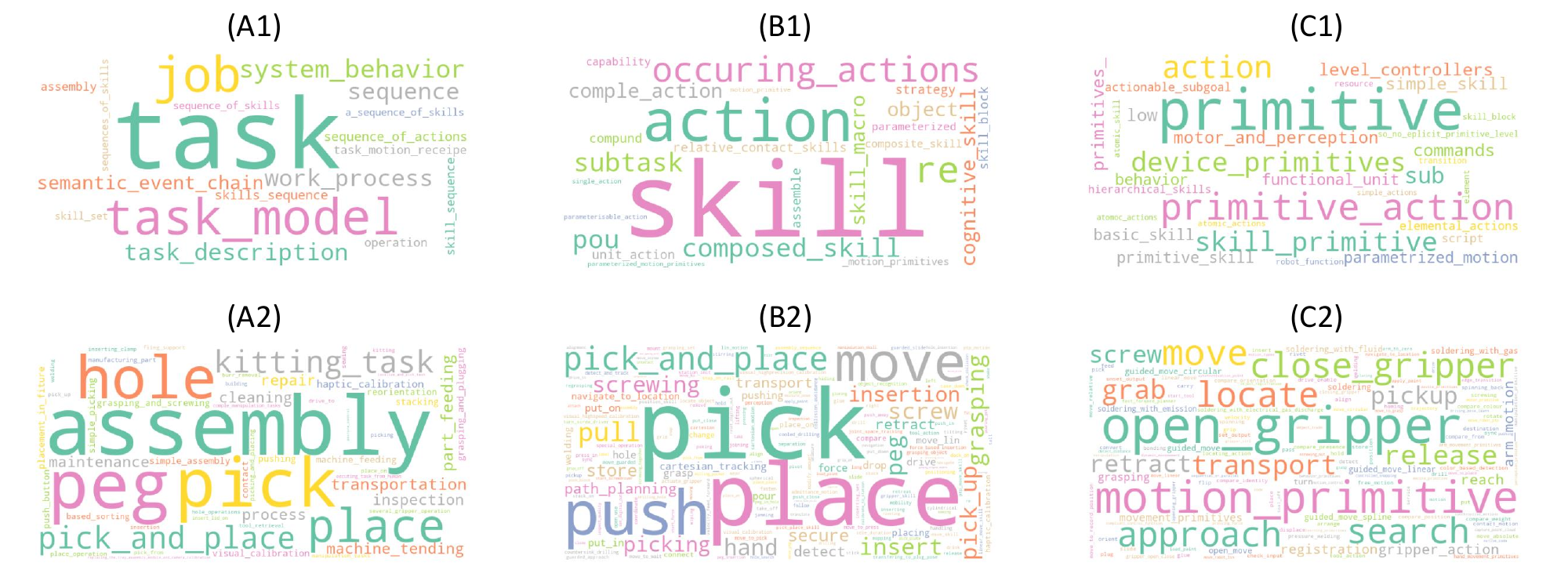}
    \caption{Wordcloud representing the occurrence of words in the classification table. On the top the names given for task (A1), skill (B1) and primitive (C1). On the bottom the most referred tasks (A2), skills (B2) and primitives (C2). The most common task was assembly, the most common skills pick and place and the most common primitives motion\_primitive and open\_gripper.}
    \label{fig:wordclouds}
\end{figure*}

A search about common terms was applied using a wordcloud\footnote{http://amueller.github.io/word\_cloud/} based on the label's frequency (bar plots are also available in the supplementary materials). The results for task, skill and primitive are visualized in Fig.~\ref{fig:wordclouds}. The naming task, skill and primitive are the most used by the research community therefore answering to the first part of \textit{RQ1}. However, other nomenclatures like action seem to be frequently used in robotics \cite{paulius2019survey}. Moreover, the most investigated types were: \emph{assembly} for task (also in line with the identified most required capability by \cite{Aaltonen.2019}), \emph{pick} and \emph{place} for skill and \emph{motion\_primitive}, and \emph{open\_gripper} for primitive.

\subsubsection{Semantic similarities with K-means clustering}
To investigate if researchers had a similar focus on action types, a K-means clustering was applied after removing duplicate terms. This section reports the cumulative analysis on primitives, skills and tasks. Initially, the terms were encoded in feature vectors using sentence transformers (SBERT) \cite{reimers2019sentencebert} with the pre-trained model \texttt{all-mpnet-base-v2}, known for its good performances in general purpose tasks. Afterwards, a K-means clustering on the encoded features was applied with the parameters of 10 clusters and dimensionality reduction to 2 for visualization purposes. Finally, for each of the obtained clusters, a keyword search was applied. The two words scoring the best cosine similarity with all the words present in that cluster were identified. %The clustering is also accompanied by the citation to the researched works allocated in the different clusters, 
The identified clusters are denoted in Table \ref{tab:reviewedworks} (visual results are also available in the supplementary material). From these results, the following insights can be drawn. Firstly, it can be perceived that the research focuses mostly on the group pick-placement, therefore answering to \textit{RQ2}. This can be related to the tasks that industrial use cases commonly face \cite{angleraud2019teaching, stenmark2014describing}.
The groups motion-movement and grip-gripper are implemented by the researchers mostly as primitives like in \cite{bolano2020virtual}, underlining that those simple capabilities are the building blocks to create different skill types. This is also reflected by the most occurring primitives (motion\_primitive, open\_gripper) identified in Sec. \ref{sec:common_terms}. The groups button-press, clean-wipe, navigate-circular, object-registration, placement-pick, rotate-spin and spray-paint are mainly implemented by the researchers in the skill level like in \cite{volkmann2020cad} and they represent several robots' capabilities necessary to accomplish tasks (i.e., clean-wipe for the task of cleaning a room). Finally, the machine-code group is integrated in the task level due to the large amount of necessary capabilities when robots have to interface with machines. For example in \cite{andersen2014definition}, the task machine feeding is identified where the robot should be capable of interacting with the machine (e.g., set inputs/outputs) and handle objects of different sizes and shapes (e.g., picking, placement, locate).

\begin{table*}[t]
  \centering
  \caption{Review table associated to the different works. The table shows the classified research works on the cluster level and on the robot's capability complexity. From the table it is easy to perceive that the placement-pick group is the most investigated by the researchers.}
  \begin{tabular}{c|p{3cm}|p{6.5cm}|p{4.4cm}}
    \hline
    \textbf{Cluster name} & \textbf{Task} & \textbf{Skill} & \textbf{Primitive}  \\
    \hline
button-press&{\tiny\cite{steinmetz2019intuitive},\cite{willibald2020collaborative},\cite{aein2013toward},\cite{stenmark2017knowledge},\cite{rovida2018motion} }&{\tiny\cite{zimmermann2019skill},\cite{calderon2010teaching},\cite{aein2013toward},\cite{pane2020skill},\cite{steinmetz2019intuitive},\cite{thomas2013new},\cite{sorensen2020towards},\cite{ji2018one},\cite{nagele2018prototype},\cite{zimmer2017bootstrapping},\cite{ahmadzadeh2015learning},\cite{zhang2016industrial},\cite{rovida2018motion},\cite{wang2018masd},\cite{dai2016robot},\cite{zhang2009autonomous},\cite{huckaby2012taxonomic},\cite{stenmark2016demonstrations},\cite{ji2021learning},\cite{rozo2020learning},\cite{yahya2017collective} }&{\tiny\cite{volkmann2020cad},\cite{ersen2017cognition},\cite{huckaby2012taxonomic},\cite{backhaus2017digital},\cite{ji2021learning},\cite{jacobsson2016modularization} }\\
\hline
clean-wipe&{\tiny\cite{bruno2018dynamic},\cite{pedersen2016robot} }&{\tiny\cite{wang2018masd},\cite{topp2018ontology},\cite{fechter2018integrated},\cite{ji2018one},\cite{calderon2010teaching},\cite{hwang2017seamless},\cite{aein2013toward},\cite{huckaby2012taxonomic},\cite{zhang2009autonomous},\cite{nagele2018prototype},\cite{kattepur2018knowledge},\cite{lindorfer2019towards},\cite{rovida2018motion},\cite{rozo2020learning} }&{\tiny\cite{volkmann2020cad},\cite{ji2021learning},\cite{stenmark2015knowledge},\cite{huckaby2012taxonomic},\cite{zimmermann2019skill} }\\
\hline
grip-gripper&{\tiny\cite{zhang2016industrial},\cite{thomas2013new},\cite{pedersen2014intuitive} }&{\tiny\cite{ji2021learning},\cite{thomas2013new},\cite{ji2018one},\cite{ahmadzadeh2015learning},\cite{cai2020inferring},\cite{fechter2018integrated},\cite{hyonyoung2020development},\cite{steinmetz2019intuitive},\cite{johannsmeier2016hierarchical},\cite{andersen2017integration},\cite{lopez2018grounding},\cite{lindorfer2019towards},\cite{kattepur2018knowledge},\cite{rozo2020learning},\cite{krueger2016vertical},\cite{calderon2010teaching},\cite{pedersen2014intuitive} }&{\tiny\cite{huckaby2012taxonomic},\cite{ji2021learning},\cite{pedersen2016robot},\cite{thomas2013new},\cite{bolano2020virtual},\cite{stenmark2014describing},\cite{calderon2010teaching},\cite{kattepur2018knowledge},\cite{krueger2019testing},\cite{rovida2017skiros},\cite{stenmark2016demonstrations},\cite{stenmark2017knowledge},\cite{andersen2016using},\cite{zimmermann2019skill},\cite{ersen2017cognition},\cite{profanter2019hardware},\cite{huckaby2012taxonomic}, \cite{pane2020skill},\cite{steinmetz2018razer} }\\
\hline
machine-code&{\tiny\cite{ersen2017cognition},\cite{bruno2018dynamic},\cite{pedersen2016robot},\cite{andersen2014definition},\cite{delhaisse2017transfer},\cite{lopez2016skill},\cite{dai2016robot},\cite{rovida2018motion},\cite{topp2018ontology},\cite{wang2018masd},\cite{krueger2019testing},\cite{krueger2016vertical},\cite{rovida2017skiros},\cite{stenmark2017simplified},\cite{andersen2017integration},\cite{volkmann2020cad} }&{\tiny\cite{johannsmeier2019framework},\cite{ersen2017cognition},\cite{kra2020production},\cite{fechter2018integrated},\cite{backhaus2017digital},\cite{nagele2018prototype},\cite{andersen2017integration},\cite{krueger2016vertical},\cite{johannsmeier2016hierarchical},\cite{lopez2016skill},\cite{stenmark2015knowledge},\cite{zhang2009autonomous},\cite{ji2021learning},\cite{heuss2020integration} }&{\tiny\cite{topp2018ontology},\cite{johannsmeier2019framework},\cite{stenmark2017simplified},\cite{krueger2019testing},\cite{rovida2017skiros},\cite{backhaus2017digital},\cite{stenmark2015knowledge},\cite{ji2021learning},\cite{calderon2010teaching},\cite{stenmark2014describing},\cite{krueger2016vertical},\cite{ersen2017cognition},\cite{andersen2014definition},\cite{stenmark2017knowledge} }\\
\hline
motion-movement&{\tiny\cite{ceriani2015reactive} }&{\tiny\cite{nagele2018prototype},\cite{yahya2017collective},\cite{pane2020skill},\cite{jha2017kinematics},\cite{huckaby2012taxonomic} }&{\tiny\cite{wang2020enhancing},\cite{ersen2017cognition},\cite{stenmark2017knowledge},\cite{dai2016robot},\cite{andersen2014definition},\cite{stenmark2015knowledge},\cite{stenmark2016demonstrations},\cite{aein2013toward},\cite{krueger2019testing},\cite{rovida2017skiros},\cite{bolano2020virtual},\cite{lankin2020ros},\cite{steinmetz2018razer},\cite{stenmark2017simplified},\cite{huang2016vision},\cite{krueger2016vertical},\cite{rovida2018motion},\cite{delhaisse2017transfer},\cite{crosby2017integrating},\cite{rovida2015design},\cite{huckaby2012taxonomic} }\\
\hline
navigate-circular&&{\tiny\cite{zhang2009autonomous},\cite{kattepur2018knowledge},\cite{profanter2019hardware},\cite{krueger2016vertical},\cite{topp2018ontology},\cite{kra2020production},\cite{huckaby2012taxonomic},\cite{backhaus2017digital},\cite{stenmark2015knowledge},\cite{hyonyoung2020development},\cite{heuss2020integration},\cite{cai2020inferring},\cite{sorensen2020towards},\cite{zhang2016industrial},\cite{nagele2018prototype},\cite{pane2020skill},\cite{wantia2016task},\cite{stenmark2015connecting},\cite{zimmermann2019skill},\cite{lankin2020ros},\cite{andersen2016using} }&{\tiny\cite{volkmann2020cad},\cite{stenmark2015knowledge},\cite{backhaus2017digital},\cite{pedersen2016robot},\cite{zimmermann2019skill},\cite{huckaby2012taxonomic},\cite{kattepur2018knowledge},\cite{stenmark2014describing},\cite{andersen2016using},\cite{johannsmeier2019framework},\cite{krueger2019testing},\cite{lankin2020ros} }\\
\hline
object-registration&{\tiny\cite{pedersen2016robot} }&{\tiny\cite{backhaus2017digital},\cite{kattepur2018knowledge},\cite{huckaby2012taxonomic},\cite{kra2020production},\cite{pedersen2016robot},\cite{wang2018masd},\cite{pane2020skill},\cite{rovida2018motion},\cite{andersen2016using},\cite{zhang2009autonomous},\cite{johannsmeier2016hierarchical},\cite{heuss2020integration} }&{\tiny\cite{rovida2017skiros},\cite{backhaus2017digital},\cite{nagele2018prototype},\cite{zhang2009autonomous},\cite{huckaby2012taxonomic},\cite{andersen2016using},\cite{andersen2014definition} }\\
\hline
placement-pick&{\tiny\cite{zhang2016industrial},\cite{andersen2014definition},\cite{aein2013toward},\cite{angleraud2019teaching},\cite{hyonyoung2020development},\cite{bolano2020virtual},\cite{andersen2016using},\cite{lindorfer2019towards},\cite{pedersen2014intuitive},\cite{stenmark2015connecting},\cite{wang2018masd} }&{\tiny\cite{zimmermann2019skill},\cite{backhaus2017digital},\cite{stenmark2015knowledge},\cite{andersen2014definition},\cite{kattepur2018knowledge},\cite{aein2013toward},\cite{steinmetz2018razer},\cite{stenmark2017simplified},\cite{rovida2015design},\cite{profanter2019hardware},\cite{heuss2020integration},\cite{crosby2017integrating},\cite{pedersen2014intuitive},\cite{andersen2017integration},\cite{zhang2009autonomous},\cite{huckaby2012taxonomic},\cite{stenmark2016demonstrations},\cite{steinmetz2019intuitive},\cite{krueger2019testing},\cite{angleraud2019teaching},\cite{willibald2020collaborative},\cite{cai2020inferring},\cite{krueger2016vertical},\cite{dai2016robot},\cite{andersen2016using},\cite{stenmark2014describing},\cite{wang2018masd},\cite{rovida2017skiros},\cite{pedersen2016robot},\cite{topp2018ontology},\cite{fechter2018integrated},\cite{nagele2018prototype},\cite{rozo2020learning},\cite{zhang2016industrial},\cite{stenmark2015connecting} }&{\tiny\cite{backhaus2017digital},\cite{zimmermann2019skill},\cite{huckaby2012taxonomic},\cite{rovida2017skiros},\cite{andersen2016using},\cite{stenmark2015knowledge},\cite{pedersen2016robot},\cite{calderon2010teaching},\cite{stenmark2014describing},\cite{stenmark2016demonstrations},\cite{kattepur2018knowledge},\cite{krueger2019testing} }\\
\hline
rotate-spin&{\tiny\cite{willibald2020collaborative},\cite{pedersen2016robot} }&{\tiny\cite{backhaus2017digital},\cite{stenmark2015knowledge},\cite{angleraud2019teaching},\cite{nagele2018prototype},\cite{delhaisse2017transfer},\cite{rozo2020learning},\cite{dai2016robot},\cite{scherzinger2019contact},\cite{huckaby2012taxonomic},\cite{stenmark2015connecting},\cite{hwang2017seamless},\cite{wantia2016task},\cite{ji2018one},\cite{cai2020inferring} }&{\tiny\cite{thomas2013new},\cite{stenmark2015knowledge},\cite{zimmermann2019skill},\cite{ji2021learning},\cite{rovida2017skiros},\cite{andersen2017integration},\cite{backhaus2017digital},\cite{huckaby2012taxonomic},\cite{stenmark2014describing},\cite{nagele2018prototype} }\\
\hline
spray-paint&{\tiny\cite{kharidege2017practical},\cite{bruno2018dynamic},\cite{schou2018skill},\cite{huang2016vision},\cite{steinmetz2019intuitive},\cite{hyonyoung2020development},\cite{cho2020learning} }&{\tiny\cite{lankin2020ros},\cite{schleipen2014automationml},\cite{volkmann2020cad},\cite{jacobsson2016modularization},\cite{lopez2016skill},\cite{cho2020learning},\cite{huckaby2012taxonomic},\cite{scherzinger2019contact},\cite{steinmetz2018razer},\cite{jha2017kinematics},\cite{dai2016robot},\cite{stenmark2015connecting},\cite{huang2016vision},\cite{steinmetz2019intuitive},\cite{ceriani2015reactive} }&{\tiny\cite{lankin2020ros},\cite{backhaus2017digital},\cite{jacobsson2016modularization},\cite{huckaby2012taxonomic} }\\
\hline
  \end{tabular}
  \label{tab:reviewedworks}
\end{table*}

\subsection{Industrial and non industrial scope}
\begin{figure}
    \centering
    \includegraphics[width=0.99\linewidth]{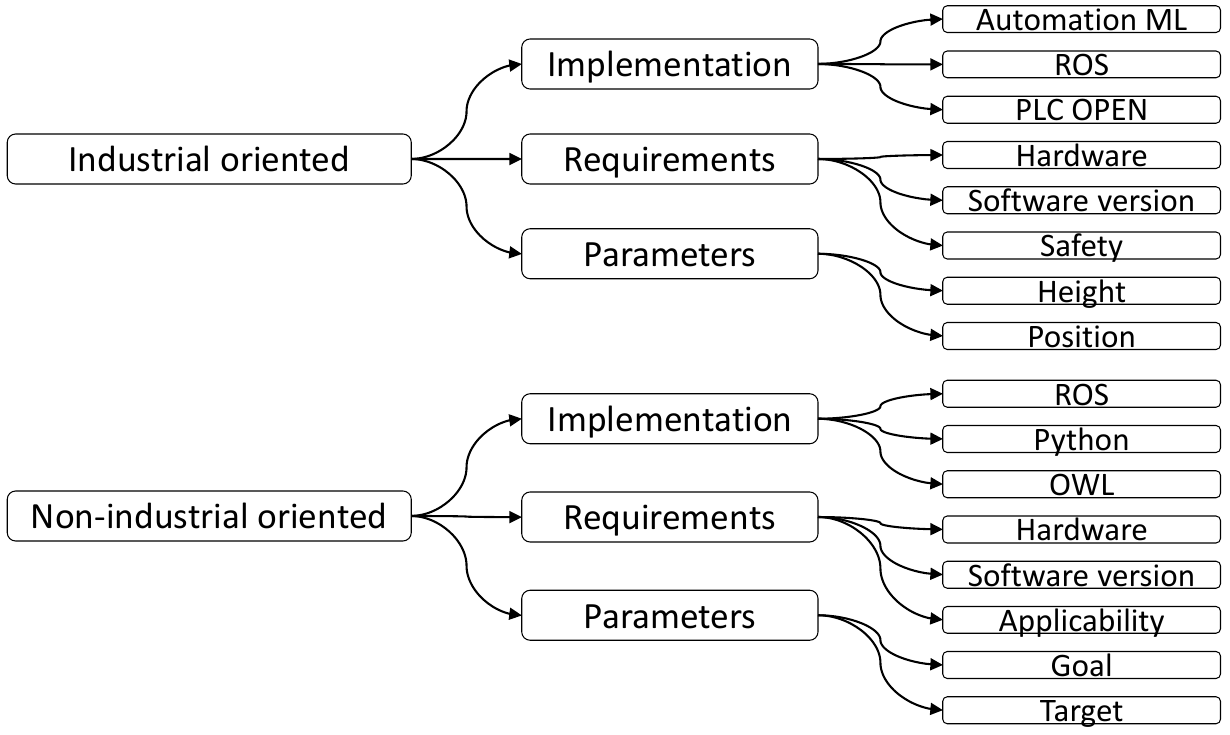}
    \caption{Diagram showing the differences in capability-based skill frameworks between works with industrial scope and non-industrial scope. The leaves on the right hand side are the most frequently appearing words.}
    \label{fig:skillReview_Diagramm}
\end{figure}

A frequency analysis was performed to identify most common terms in the two sub-sets given by filtering the \textit{Industrial} criteria for identifying data regarding \textit{RQ3}. The overall analysis can be seen in Fig.~\ref{fig:skillReview_Diagramm}.

\subsubsection{Implementations}
Industrial scenarios show a diverse set of frameworks closer to the automation domain (i.e. programmable logic controller (PLC) language, AutomationML (AML)) \cite{backhaus2017digital, zimmermann2019skill}. Additionally, the Robot Operating System (ROS) also finds its way into such scenarios \cite{pedersen2016robot, akkaladevi2019skill}. In comparison, non-industrial applications rely quite often on ROS with the python programming language, such as in \cite{bolano2020virtual}. Hereby, ROS has the main purpose to serve as a communication middleware between so-called nodes but it also defines interfaces in the form of standardized message formats. However, ROS is not used to implement knowledge itself, which is an important requirement for non-industrial applications. Thus, some works rely on ontological representations. Ontologies can be implemented in the W3C Web Ontology Language (OWL), and some of them were already standardized, such as the \emph{IEEE 1872} Core Ontology for Robotics and Automation (CORA) first proposed in \cite{schlenoff2012ieee} and validated in \cite{de2019prototyping}. 

\subsubsection{Industrial requirements}
The requirements regarding type of hardware used, software version and robot intended behaviour were equally considered both for the industrial and the non-industrial scenarios like in \cite{schleipen2014automationml, krueger2016vertical}. The major difference is that some of the industrial scenarios consider also the requirements related to safety aspects of the application like in \cite{nagele2018prototype, ceriani2015reactive}. This has been always a major point of difference between industrial and non-industrial research \cite{Aaltonen.2019}, and this is reflected also in this review concerning skill frameworks. Therefore, to increase market adoption of skill frameworks in the industrial domain, safety should be addressed either by having inherently safe skills or by conducting a risk analysis behind each robotic skill.

\subsubsection{Parameters}
In the industrially focused works, the parameter scope of a skill is closer to hardware functions and physically measurable data. The most used parameters were height, position, offset, and robot\_speed. These parameters appear to allow only minor adaptions to the robotic task and appear to take rather hardcoded values such as the absolute position of an object provided by the programmer \cite{pedersen2016robot, akkaladevi2019skill}. In comparison, the complexity of parameters in the non-industrial scenarios is considered to be higher. The most commonly used parameters were goal, target, world, and robot. Parameters of this scope allow major modifications of the skill's behavior. A skill that is parameterized with an abstract target instead of an absolute position could adapt its behavior significantly by exploiting further information from a knowledge base like in \cite{huckaby2012taxonomic, pane2020skill}. Parameters can be also seen as function arguments that can be either passed to a skill or a primitive. Considering different programming layers, parameters would describe the input ports of a skill visible to the user, while parameters could also describe the function arguments of a primitive, which are visible only to the system designer. In both industrial and non-industrial cases, the parameters were mostly found to be associated with a skill. The internal logic of the skill is then meant to extract meaningful values that are shared with the underlying primitives. Examples of this structure can be also found in~\cite{dai2016robot, profanter2019hardware}.\newline

With these findings it is possible to find the answers to \textit{RQ3}. Industrial usage of robot capabilities distinguishes itself from non-industrial usages on two areas. First, its focus is on simple skills with often hard-coded parameters. Second, the implementation uses frameworks close to the automation domain while always respecting the safety of the application.

\subsection{Approaches to capability-based skill frameworks}
Finally, from the review it was possible to see that 57 out of 88 papers used a similar architecture as the one proposed here. Therefore, a tree-like structure where primitives are the closest units to the hardware level and the tasks the farther away from the hardware seems to be a concept that most researchers agree on, both for the industrial and the non-industrial cases. Therefore, answering to the second part of \textit{RQ1}. The best examples on the usage of the identified architecture can be found in \cite{pedersen2016robot, rovida2017skiros, profanter2019hardware}. In these works, also a common skill model is presented. Often this model is dependent on the resource which provides certain functionalities (i.e., primitive) and the input/output variables which can parameterize the functionality. The best example of such modelling can be seen in \cite{backhaus2017digital}. 

\section{Trends and outlooks \label{sec:discussion}}
During the review, the necessity to accommodate market demands leading to HMLV productions has been underlined as also identified by \cite{Froschauer.4262022}. To adapt automation in such production scenarios, robots with skill frameworks were seen as enabling technology within Industrie 4.0 \cite{verhoosel2017recipe, johansen2021role}. The aim of this technology is to avoid the high costs of manual processes on the one hand and the limitations of fully automatic, poorly customizable processes on the other hand. To properly exploit the advantages of skill frameworks, however, skill hardware and vendor independence is a key factor as long as it guarantees wide skill applicability \cite{Schleipen.2015} and, for example, skills could be used across multiple plant sites of manufacturing companies \cite{jacobsson2016modularization}. To enable such independence, primitives will need to be properly mapped to skills according to the available hardware functionalities. Therefore, an automatic primitive to skill mapping is worth investigating \cite{andersen2017integration}. However, such mapping would require an universal information representation among all employed skills \cite{topp2018ontology}. To lay a foundation for that, skills and their primitives could be defined in industrial standards such as AML like in \cite{backhaus2017digital} or definitions like the FoF ontology \cite{schafer2021flexible}. Within this aspect, it is worth noting that industrial applications initially preferred AML as information representation and this is also visible in Fig.~\ref{fig:skillReview_Diagramm}. However, in the last years, OPC UA has become more common \cite{profanter2019hardware} and none of the surveyed works report to use AML in the last three years. Apart from skill definitions, parameters are important to enable skills' reusability. In many industrial applications, skill parameters are still manually defined \cite{nagele2018prototype}. However, recent works consider automatic parameterization techniques, where the skill sequence and skill parameters are defined either by an autonomous planner or extracted from human demonstrations \cite{willibald2020collaborative, steinmetz2019intuitive}. Also, the complexity of parameters is changing. From simple, physical quantities such as positions, parameters are moving towards more abstract ones, such as object IDs or even interfaces to world models which are passed as parameters \cite{backhaus2017digital}. This shows that the responsibility of interpreting a parameter is being shifted from the human to the skill itself.

\iffalse
\begin{figure}
    \centering
     \includegraphics[clip,width=0.95\linewidth]{example-image-a}
    \caption{Conceptual model of a skill by \cite{backhaus2017digital}. A skill is dependent on the resource which provides certain functionalities (i.e., primitive) and the input/output variables which can parameterize the functionality.}
    \label{fig:skill_models.backhaus}
\end{figure}
\fi

\iffalse
\begin{figure}
    \centering
    \includegraphics[width=0.95\linewidth]{figures/automationTrend.pdf}
    \caption{Processes and involved technologies upon batch size and product variants. Robots coupled with skills are seen as the biggest enabler for flexible processes necessary to accommodate HMLV.}
    \label{fig:automationTrend}
\end{figure}
\fi

\section{Conclusions \label{sec:conclusion}}
This work presented the review of several papers on the field of capability-based skill frameworks by focusing on the robotic industrial domain. The review was performed via a structured approach and the research works were classified according to some predefined criteria. The results obtained were then analyzed using semantic clustering or frequency of appearance. Through this methodology, the following results were obtained. Firstly, the analysis showed that the research practitioners, when referring to capabilities, use often the names task, skill and primitive, where primitive is the closest to hardware, tasks the furthest and skills represent robots' capabilities. Secondly, several research areas defined by the type of robotic capability have been identified. From this classification it was discovered that pick-placement is the most researched capability and that motion-movement, grip-gripper are common groups of primitives used to create different skills. Finally, some differences have been found between industrial (\emph{I}) and non-industrial (\emph{NI}) research. \emph{I} uses parameters that are close to the hardware, whereas \emph{NI} uses high level ones. Considering implementation frameworks, \emph{I} prefers PLC languages while \emph{NI} others (e.g., ROS). For industrial requirements, it was found that \emph{I} considers more the safety aspects when compared to \emph{NI}. Apart from these main findings, the review also showed an increasing interest on the usage of robots and skills to accommodate requirements of a HMLV production and that information representation is essential to enable skill reusability, either via OPC UA or other standards. 
%With this review, the robotic research community is directed to the most investigated robots' capabilities in the industrial domain. 
While performing the review, a major pitfall was identified on the research query. Such query was largely biased towards industrial scenarios and might better represent this domain compared to non-industrial one. Therefore, future work could focus more on the review of purely academic works in robots' capabilities to give a bigger picture of the field.
\bibliographystyle{IEEEtran}
\bibliography{bibliography.bib}

\begin{thebibliography}{10}
\providecommand{\url}[1]{#1}
\csname url@rmstyle\endcsname
\providecommand{\newblock}{\relax}
\providecommand{\bibinfo}[2]{#2}
\providecommand\BIBentrySTDinterwordspacing{\spaceskip=0pt\relax}
\providecommand\BIBentryALTinterwordstretchfactor{4}
\providecommand\BIBentryALTinterwordspacing{\spaceskip=\fontdimen2\font plus
\BIBentryALTinterwordstretchfactor\fontdimen3\font minus
  \fontdimen4\font\relax}
\providecommand\BIBforeignlanguage[2]{{%
\expandafter\ifx\csname l@#1\endcsname\relax
\typeout{** WARNING: IEEEtran.bst: No hyphenation pattern has been}%
\typeout{** loaded for the language `#1'. Using the pattern for}%
\typeout{** the default language instead.}%
\else
\language=\csname l@#1\endcsname
\fi
#2}}

\bibitem{Anshuk.2014}
G.~Anshuk, C.~Magar, and R.~Roger, ``How technology can drive the next wave of
  mass customization: Seven technologies are making it easier to tailor
  products and services to the wants of individual customers---and still make a
  profit.'' 2014.

\bibitem{Fechter.2018}
M.~Fechter, C.~Seeber, and S.~Chen, ``Integrated process planning and resource
  allocation for collaborative robot workplace design,'' \emph{Procedia CIRP},
  vol.~72, pp. 39--44, 2018.

\bibitem{bayha2020describing}
A.~Bayha, J.~Bock, B.~Boss, C.~Diedrich, and S.~Malakuti, ``Describing
  capabilities of industrie 4.0 components,'' [online] Available:
  \url{https://www.plattform-i40.de/IP/Redaktion/EN/Downloads/Publikation/Capabilities_Industrie40_Components.html}
  (accessed on 2022/04/24), p.~33, 2020.

\bibitem{CuttingDecelle.2007}
A.~F. Cutting-Decelle, R.~Young, J.~J. Michel, R.~Grangel, J.~{Le Cardinal},
  and J.~P. Bourey, ``Iso 15531 mandate: A product-process-resource based
  approach for managing modularity in production management,'' \emph{Concurrent
  Engineering}, vol.~15, no.~2, pp. 217--235, 2007.

\bibitem{backhaus2017digital}
J.~Backhaus and G.~Reinhart, ``Digital description of products, processes and
  resources for task-oriented programming of assembly systems,'' \emph{Journal
  of Intelligent Manufacturing}, vol.~28, no.~8, pp. 1787--1800, 2017.

\bibitem{Froschauer.4262022}
R.~Froschauer, A.~K{\"o}cher, K.~Meixner, S.~Schmitt, and F.~Spitzer,
  ``Capabilities and skills in manufacturing: A survey over the last decade of
  etfa,'' [online] Available: \url{http://arxiv.org/pdf/2204.12908v1} (accessed
  on 2022/06/21).

\bibitem{Booth.2021}
A.~Booth, A.~Sutton, M.~Clowes, and M.~{Martyn-St James}, \emph{{Systematic
  approaches to a successful literature review}}, third edition / andrew booth,
  anthea sutton, mark clowes, marrissa martyn-st james~ed.\hskip 1em plus 0.5em
  minus 0.4em\relax Los Angeles: SAGE, 2021.

\bibitem{Schleipen.2015}
M.~Schleipen, A.~L{\"u}der, O.~Sauer, H.~Flatt, and J.~Jasperneite,
  ``{Requirements and concept for Plug-and-Work},'' \emph{{at -
  Automatisierungstechnik}}, vol.~63, no.~10, pp. 801--820, 2015.

\bibitem{schafer2021flexible}
P.~M. Sch{\"a}fer, F.~Steinmetz, S.~Schneyer, T.~Bachmann, T.~Eiband, F.~S.
  Lay, A.~Padalkar, C.~S{\"u}rig, F.~Stulp, and K.~Nottensteiner, ``Flexible
  robotic assembly based on ontological representation of tasks, skills, and
  resources,'' in \emph{Proceedings of the International Conference on
  Principles of Knowledge Representation and Reasoning}, vol.~18, no.~1, 2021,
  pp. 702--706.

\bibitem{andersen2014definition}
R.~H. Andersen, T.~Solund, and J.~Hallam, ``Definition and initial case-based
  evaluation of hardware-independent robot skills for industrial robotic
  co-workers,'' in \emph{ISR/Robotik 2014; 41st International Symposium on
  Robotics}.\hskip 1em plus 0.5em minus 0.4em\relax VDE, 2014, pp. 1--7.

\bibitem{huckaby2012taxonomic}
J.~O. Huckaby and H.~I. Christensen, ``A taxonomic framework for task modeling
  and knowledge transfer in manufacturing robotics,'' in \emph{Workshops at the
  Twenty-Sixth AAAI Conference on Artificial Intelligence}, 2012.

\bibitem{ji2021learning}
S.~Ji, S.~Lee, S.~Yoo, I.~Suh, I.~Kwon, F.~C. Park, S.~Lee, and H.~Kim,
  ``Learning-based automation of robotic assembly for smart manufacturing,''
  \emph{Proceedings of the IEEE}, vol. 109, no.~4, pp. 423--440, 2021.

\bibitem{InternationalElectrotechnicalCommission.2013}
{International Electrotechnical Commission}, ``{IEC 62264 - Enterprise-control
  system integration},'' 2013.

\bibitem{pedersen2016robot}
M.~R. Pedersen, L.~Nalpantidis, R.~S. Andersen, C.~Schou, S.~B{\o}gh,
  V.~Kr{\"u}ger, and O.~Madsen, ``Robot skills for manufacturing: From concept
  to industrial deployment,'' \emph{Robotics and Computer-Integrated
  Manufacturing}, vol.~37, pp. 282--291, 2016.

\bibitem{profanter2019hardware}
S.~Profanter, A.~Breitkreuz, M.~Rickert, and A.~Knoll, ``A hardware-agnostic
  opc ua skill model for robot manipulators and tools,'' in \emph{24th IEEE
  International Conference on Emerging Technologies and Factory Automation
  (ETFA)}, 2019, pp. 1061--1068.

\bibitem{steinmetz2018razer}
F.~Steinmetz, A.~Wollschl{\"a}ger, and R.~Weitschat, ``Razer—a hri for visual
  task-level programming and intuitive skill parameterization,'' \emph{IEEE
  Robotics and Automation Letters}, vol.~3, no.~3, pp. 1362--1369, 2018.

\bibitem{R.H.Andersen.2014}
{R. H. Andersen}, {T. Solund}, and {J. Hallam}, ``{Definition and Initial
  Case-Based Evaluation of Hardware-Independent Robot Skills for Industrial
  Robotic Co-Workers},'' in \emph{{ISR/Robotik 2014; 41st International
  Symposium on Robotics}}, 2014, pp. 1--7.

\bibitem{Schaal.2006}
S.~Schaal, ``{Dynamic Movement Primitives -A Framework for Motor Control in
  Humans and Humanoid Robotics},'' in \emph{{Adaptive Motion of Animals and
  Machines}}, H.~Kimura, K.~Tsuchiya, A.~Ishiguro, and H.~Witte, Eds.\hskip 1em
  plus 0.5em minus 0.4em\relax Tokyo: Springer-Verlag, 2006, pp. 261--280.

\bibitem{loper2002nltk}
E.~Loper and S.~Bird, ``Nltk: The natural language toolkit,'' \emph{arXiv
  preprint cs/0205028}, 2002.

\bibitem{paulius2019survey}
D.~Paulius and Y.~Sun, ``A survey of knowledge representation in service
  robotics,'' \emph{Robotics and Autonomous Systems}, vol. 118, pp. 13--30,
  2019.

\bibitem{Aaltonen.2019}
I.~Aaltonen and T.~Salmi, ``{Experiences and expectations of collaborative
  robots in industry and academia: barriers and development needs},''
  \emph{{Procedia Manufacturing}}, vol.~38, pp. 1151--1158, 2019.

\bibitem{reimers2019sentencebert}
N.~Reimers and I.~Gurevych, ``Sentence-bert: Sentence embeddings using siamese
  bert-networks,'' 2019.

\bibitem{angleraud2019teaching}
A.~Angleraud, Q.~Houbre, and R.~Pieters, ``Teaching semantics and skills for
  human-robot collaboration,'' \emph{Paladyn, Journal of Behavioral Robotics},
  vol.~10, no.~1, pp. 318--329, 2019.

\bibitem{stenmark2014describing}
M.~Stenmark and J.~Malec, ``Describing constraint-based assembly tasks in
  unstructured natural language,'' \emph{IFAC Proceedings Volumes}, vol.~47,
  no.~3, pp. 3056--3061, 2014.

\bibitem{bolano2020virtual}
G.~Bolano, A.~Roennau, R.~Dillmann, and A.~Groz, ``Virtual reality for offline
  programming of robotic applications with online teaching methods,'' in
  \emph{17th International Conference on Ubiquitous Robots (UR)}.\hskip 1em
  plus 0.5em minus 0.4em\relax IEEE, 2020, pp. 625--630.

\bibitem{volkmann2020cad}
M.~Volkmann, T.~Legler, A.~Wagner, and M.~Ruskowski, ``A cad feature-based
  manufacturing approach with opc ua skills,'' \emph{Procedia Manufacturing},
  vol.~51, pp. 416--423, 2020.

\bibitem{steinmetz2019intuitive}
F.~Steinmetz, V.~Nitsch, and F.~Stulp, ``Intuitive task-level programming by
  demonstration through semantic skill recognition,'' \emph{IEEE Robotics and
  Automation Letters}, vol.~4, no.~4, pp. 3742--3749, 2019.

\bibitem{willibald2020collaborative}
C.~Willibald, T.~Eiband, and D.~Lee, ``Collaborative programming of conditional
  robot tasks,'' in \emph{IEEE/RSJ International Conference on Intelligent
  Robots and Systems (IROS)}, 2020, pp. 5402--5409.

\bibitem{aein2013toward}
M.~J. Aein, E.~E. Aksoy, M.~Tamosiunaite, J.~Papon, A.~Ude, and
  F.~W{\"o}rg{\"o}tter, ``Toward a library of manipulation actions based on
  semantic object-action relations,'' in \emph{IEEE/RSJ International
  Conference on Intelligent Robots and Systems}, 2013, pp. 4555--4562.

\bibitem{stenmark2017knowledge}
M.~Stenmark, E.~A. Topp, M.~Haage, and J.~Malec, ``Knowledge for synchronized
  dual-arm robot programming,'' in \emph{2017 AAAI Fall Symposium Series},
  2017.

\bibitem{rovida2018motion}
F.~Rovida, D.~Wuthier, B.~Grossmann, M.~Fumagalli, and V.~Kr{\"u}ger, ``Motion
  generators combined with behavior trees: a novel approach to skill
  modelling,'' in \emph{IEEE/RSJ International Conference on Intelligent Robots
  and Systems (IROS)}, 2018, pp. 5964--5971.

\bibitem{zimmermann2019skill}
P.~Zimmermann, E.~Axmann, B.~Brandenbourger, K.~Dorofeev, A.~Mankowski, and
  P.~Zanini, ``Skill-based engineering and control on field-device-level with
  opc ua,'' in \emph{24th IEEE International Conference on Emerging
  Technologies and Factory Automation (ETFA)}, 2019, pp. 1101--1108.

\bibitem{calderon2010teaching}
C.~A.~A. Calderon, R.~E. Mohan, and C.~Zhou, ``Teaching new tricks to a robot
  learning to solve a task by imitation,'' in \emph{IEEE Conference on
  Robotics, Automation and Mechatronics}, 2010, pp. 256--262.

\bibitem{pane2020skill}
Y.~Pane, E.~Aertbeli{\"e}n, J.~De~Schutter, and W.~Decr{\'e}, ``Skill-based
  programming framework for composable reactive robot behaviors,'' in
  \emph{IEEE/RSJ International Conference on Intelligent Robots and Systems
  (IROS)}, 2020, pp. 7087--7094.

\bibitem{thomas2013new}
U.~Thomas, G.~Hirzinger, B.~Rumpe, C.~Schulze, and A.~Wortmann, ``A new skill
  based robot programming language using uml/p statecharts,'' in \emph{IEEE
  International Conference on Robotics and Automation}, 2013, pp. 461--466.

\bibitem{sorensen2020towards}
L.~C. S{\o}rensen, S.~Mathiesen, R.~Waspe, and C.~Schlette, ``Towards digital
  twins for industrial assembly-improving robot solutions by intuitive user
  guidance and robot programming,'' in \emph{25th IEEE International Conference
  on Emerging Technologies and Factory Automation (ETFA)}, vol.~1, 2020, pp.
  1480--1484.

\bibitem{ji2018one}
Y.~Ji, Y.~Yang, X.~Xu, and H.~T. Shen, ``One-shot learning based pattern
  transition map for action early recognition,'' \emph{Signal Processing}, vol.
  143, pp. 364--370, 2018.

\bibitem{nagele2018prototype}
F.~N{\"a}gele, L.~Halt, P.~Tenbrock, and A.~Pott, ``A prototype-based skill
  model for specifying robotic assembly tasks,'' in \emph{IEEE International
  Conference on Robotics and Automation (ICRA)}, 2018, pp. 558--565.

\bibitem{zimmer2017bootstrapping}
M.~Zimmer and S.~Doncieux, ``Bootstrapping $ q $-learning for robotics from
  neuro-evolution results,'' \emph{IEEE Transactions on Cognitive and
  Developmental Systems}, vol.~10, no.~1, pp. 102--119, 2017.

\bibitem{ahmadzadeh2015learning}
S.~R. Ahmadzadeh, A.~Paikan, F.~Mastrogiovanni, L.~Natale, P.~Kormushev, and
  D.~G. Caldwell, ``Learning symbolic representations of actions from human
  demonstrations,'' in \emph{IEEE International Conference on Robotics and
  Automation (ICRA)}, 2015, pp. 3801--3808.

\bibitem{zhang2016industrial}
J.~Zhang, Y.~Wang, and R.~Xiong, ``Industrial robot programming by
  demonstration,'' in \emph{IEEE International Conference on Advanced Robotics
  and Mechatronics (ICARM)}, 2016, pp. 300--305.

\bibitem{wang2018masd}
Y.~Wang, Y.~Jiao, R.~Xiong, H.~Yu, J.~Zhang, and Y.~Liu, ``Masd: A multimodal
  assembly skill decoding system for robot programming by demonstration,''
  \emph{IEEE Transactions on Automation Science and Engineering}, vol.~15,
  no.~4, pp. 1722--1734, 2018.

\bibitem{dai2016robot}
F.~Dai, A.~Wahrburg, B.~Matthias, and H.~Ding, ``Robot assembly skills based on
  compliant motion,'' in \emph{Proceedings of ISR 2016: 47st International
  Symposium on Robotics}.\hskip 1em plus 0.5em minus 0.4em\relax VDE, 2016, pp.
  1--6.

\bibitem{zhang2009autonomous}
J.~Zhang \emph{et~al.}, ``Autonomous planning for mobile manipulation services
  based on multi-level robot skills,'' in \emph{IEEE/RSJ International
  Conference on Intelligent Robots and Systems}, 2009, pp. 1999--2004.

\bibitem{stenmark2016demonstrations}
M.~Stenmark and E.~A. Topp, ``From demonstrations to skills for high-level
  programming of industrial robots,'' in \emph{2016 AAAI fall symposium
  series}, 2016.

\bibitem{rozo2020learning}
L.~Rozo, M.~Guo, A.~G. Kupcsik, M.~Todescato, P.~Schillinger, M.~Giftthaler,
  M.~Ochs, M.~Spies, N.~Waniek, P.~Kesper, \emph{et~al.}, ``Learning and
  sequencing of object-centric manipulation skills for industrial tasks,'' in
  \emph{IEEE/RSJ International Conference on Intelligent Robots and Systems
  (IROS)}, 2020, pp. 9072--9079.

\bibitem{yahya2017collective}
A.~Yahya, A.~Li, M.~Kalakrishnan, Y.~Chebotar, and S.~Levine, ``Collective
  robot reinforcement learning with distributed asynchronous guided policy
  search,'' in \emph{IEEE/RSJ International Conference on Intelligent Robots
  and Systems (IROS)}, 2017, pp. 79--86.

\bibitem{ersen2017cognition}
M.~Ersen, E.~Oztop, and S.~Sariel, ``Cognition-enabled robot manipulation in
  human environments: requirements, recent work, and open problems,''
  \emph{IEEE Robotics \& Automation Magazine}, vol.~24, no.~3, pp. 108--122,
  2017.

\bibitem{jacobsson2016modularization}
L.~Jacobsson, J.~Malec, and K.~Nilsson, ``Modularization of skill ontologies
  for industrial robots,'' in \emph{Proceedings of ISR 2016: 47st international
  symposium on robotics}.\hskip 1em plus 0.5em minus 0.4em\relax VDE, 2016, pp.
  1--6.

\bibitem{bruno2018dynamic}
G.~Bruno and D.~Antonelli, ``Dynamic task classification and assignment for the
  management of human-robot collaborative teams in workcells,'' \emph{The
  International Journal of Advanced Manufacturing Technology}, vol.~98, no.~9,
  pp. 2415--2427, 2018.

\bibitem{topp2018ontology}
E.~A. Topp, M.~Stenmark, A.~Ganslandt, A.~Svensson, M.~Haage, and J.~Malec,
  ``Ontology-based knowledge representation for increased skill reusability in
  industrial robots,'' in \emph{IEEE/RSJ International Conference on
  Intelligent Robots and Systems (IROS)}, 2018, pp. 5672--5678.

\bibitem{fechter2018integrated}
M.~Fechter, C.~Seeber, and S.~Chen, ``Integrated process planning and resource
  allocation for collaborative robot workplace design,'' \emph{Procedia CIRP},
  vol.~72, pp. 39--44, 2018.

\bibitem{hwang2017seamless}
J.~Hwang and J.~Tani, ``Seamless integration and coordination of cognitive
  skills in humanoid robots: A deep learning approach,'' \emph{IEEE
  Transactions on Cognitive and Developmental Systems}, vol.~10, no.~2, pp.
  345--358, 2017.

\bibitem{kattepur2018knowledge}
A.~Kattepur, S.~Dey, and P.~Balamuralidhar, ``Knowledge based hierarchical
  decomposition of industry 4.0 robotic automation tasks,'' in \emph{IECON 44th
  Annual Conference of the IEEE Industrial Electronics Society}, 2018, pp.
  3665--3672.

\bibitem{lindorfer2019towards}
R.~Lindorfer and R.~Froschauer, ``Towards user-oriented programming of
  skill-based automation systems using a domain-specific meta-modeling
  approach,'' in \emph{IEEE 17th International Conference on Industrial
  Informatics (INDIN)}, vol.~1, 2019, pp. 655--660.

\bibitem{stenmark2015knowledge}
M.~Stenmark and J.~Malec, ``Knowledge-based instruction of manipulation tasks
  for industrial robotics,'' \emph{Robotics and Computer-Integrated
  Manufacturing}, vol.~33, pp. 56--67, 2015.

\bibitem{pedersen2014intuitive}
M.~R. Pedersen, D.~L. Herzog, and V.~Kr{\"u}ger, ``Intuitive skill-level
  programming of industrial handling tasks on a mobile manipulator,'' in
  \emph{IEEE/RSJ International Conference on Intelligent Robots and Systems},
  2014, pp. 4523--4530.

\bibitem{cai2020inferring}
C.~Cai, Y.~S. Liang, N.~Somani, and W.~Yan, ``Inferring the geometric nullspace
  of robot skills from human demonstrations,'' in \emph{IEEE International
  Conference on Robotics and Automation (ICRA)}, 2020, pp. 7668--7675.

\bibitem{hyonyoung2020development}
H.~Hyonyoung, L.~Eunseo, and K.~Hyunchul, ``Development of unit robot skill
  based task recipe for task planning,'' in \emph{2020 International Conference
  on Information and Communication Technology Convergence (ICTC)}.\hskip 1em
  plus 0.5em minus 0.4em\relax IEEE, 2020, pp. 1720--1722.

\bibitem{johannsmeier2016hierarchical}
L.~Johannsmeier and S.~Haddadin, ``A hierarchical human-robot
  interaction-planning framework for task allocation in collaborative
  industrial assembly processes,'' \emph{IEEE Robotics and Automation Letters},
  vol.~2, no.~1, pp. 41--48, 2016.

\bibitem{andersen2017integration}
R.~E. Andersen, E.~B. Hansen, D.~Cerny, S.~Madsen, B.~Pulendralingam,
  S.~B{\o}gh, and D.~Chrysostomou, ``Integration of a skill-based collaborative
  mobile robot in a smart cyber-physical environment,'' \emph{Procedia
  Manufacturing}, vol.~11, pp. 114--123, 2017.

\bibitem{lopez2018grounding}
I.~Lopez-Juarez, R.~Rios-Cabrera, E.~Rojas-Sanchez, A.~Maldonado-Ramirez, and
  G.~Lefranc, ``Grounding the lexicon for human-robot interaction during the
  manipulation of irregular objects,'' in \emph{7th International Conference on
  Computers Communications and Control (ICCCC)}, 2018, pp. 282--288.

\bibitem{krueger2016vertical}
V.~Krueger, A.~Chazoule, M.~Crosby, A.~Lasnier, M.~R. Pedersen, F.~Rovida,
  L.~Nalpantidis, R.~Petrick, C.~Toscano, and G.~Veiga, ``A vertical and
  cyber--physical integration of cognitive robots in manufacturing,''
  \emph{Proceedings of the IEEE}, vol. 104, no.~5, pp. 1114--1127, 2016.

\bibitem{krueger2019testing}
V.~Krueger, F.~Rovida, B.~Grossmann, R.~Petrick, M.~Crosby, A.~Charzoule, G.~M.
  Garcia, S.~Behnke, C.~Toscano, and G.~Veiga, ``Testing the vertical and
  cyber-physical integration of cognitive robots in manufacturing,''
  \emph{Robotics and computer-integrated manufacturing}, vol.~57, pp. 213--229,
  2019.

\bibitem{rovida2017skiros}
F.~Rovida, M.~Crosby, D.~Holz, A.~S. Polydoros, B.~Gro{\ss}mann, R.~Petrick,
  and V.~Kr{\"u}ger, ``Skiros—a skill-based robot control platform on top of
  ros,'' in \emph{Robot operating system (ROS)}.\hskip 1em plus 0.5em minus
  0.4em\relax Springer, 2017, pp. 121--160.

\bibitem{andersen2016using}
R.~S. Andersen, C.~Schou, J.~S. Damgaard, and O.~Madsen, ``Using a flexible
  skill-based approach to recognize objects in industrial scenarios,'' in
  \emph{Proceedings of ISR 2016: 47st International Symposium on
  Robotics}.\hskip 1em plus 0.5em minus 0.4em\relax VDE, 2016, pp. 1--8.

\bibitem{delhaisse2017transfer}
B.~Delhaisse, D.~Esteban, L.~Rozo, and D.~Caldwell, ``Transfer learning of
  shared latent spaces between robots with similar kinematic structure,'' in
  \emph{International Joint Conference on Neural Networks (IJCNN)}.\hskip 1em
  plus 0.5em minus 0.4em\relax IEEE, 2017, pp. 4142--4149.

\bibitem{lopez2016skill}
I.~Lopez-Juarez, ``Skill acquisition for industrial robots: From stand-alone to
  distributed learning,'' in \emph{IEEE International Conference on Automatica
  (ICA-ACCA)}, 2016, pp. 1--5.

\bibitem{stenmark2017simplified}
M.~Stenmark, M.~Haage, and E.~A. Topp, ``Simplified programming of re-usable
  skills on a safe industrial robot: Prototype and evaluation,'' in
  \emph{Proceedings of the 2017 ACM/IEEE International Conference on
  Human-Robot Interaction}, 2017, pp. 463--472.

\bibitem{johannsmeier2019framework}
L.~Johannsmeier, M.~Gerchow, and S.~Haddadin, ``A framework for robot
  manipulation: Skill formalism, meta learning and adaptive control,'' in
  \emph{IEEE International Conference on Robotics and Automation (ICRA)}, 2019,
  pp. 5844--5850.

\bibitem{kra2020production}
M.~Kr{\"a}, L.~Vogt, C.~H{\"a}rdtlein, S.~Schiele, and J.~Schilp, ``Production
  planning for collaborating resources in cyber-physical production systems,''
  \emph{Procedia CIRP}, vol.~93, pp. 192--197, 2020.

\bibitem{heuss2020integration}
L.~Heuss and G.~Reinhart, ``Integration of autonomous task planning into
  reconfigurable skill-based industrial robots,'' in \emph{25th IEEE
  International Conference on Emerging Technologies and Factory Automation
  (ETFA)}, vol.~1, 2020, pp. 1293--1296.

\bibitem{ceriani2015reactive}
N.~M. Ceriani, A.~M. Zanchettin, P.~Rocco, A.~Stolt, and A.~Robertsson,
  ``Reactive task adaptation based on hierarchical constraints classification
  for safe industrial robots,'' \emph{IEEE/ASME Transactions on Mechatronics},
  vol.~20, no.~6, pp. 2935--2949, 2015.

\bibitem{jha2017kinematics}
A.~Jha, S.~S. Chiddarwar, V.~Alakshendra, and M.~V. Andulkar,
  ``Kinematics-based approach for robot programming via human arm motion,''
  \emph{Journal of the Brazilian Society of Mechanical Sciences and
  Engineering}, vol.~39, no.~7, pp. 2659--2675, 2017.

\bibitem{wang2020enhancing}
L.~Wang, S.~Jia, G.~Wang, A.~Turner, and S.~Ratchev, ``Enhancing learning
  capabilities of movement primitives under distributed probabilistic framework
  for assembly tasks,'' in \emph{IEEE International Conference on Systems, Man,
  and Cybernetics (SMC)}, 2020, pp. 3832--3838.

\bibitem{lankin2020ros}
R.~Lankin, K.~Kim, and P.-C. Huang, ``Ros-based robot simulation for repetitive
  labor-intensive construction tasks,'' in \emph{IEEE 18th International
  Conference on Industrial Informatics (INDIN)}, vol.~1, 2020, pp. 206--213.

\bibitem{huang2016vision}
B.~Huang, A.~Vandini, Y.~Hu, S.-L. Lee, and G.-Z. Yang, ``A vision-guided dual
  arm sewing system for stent graft manufacturing,'' in \emph{IEEE/RSJ
  International Conference on Intelligent Robots and Systems (IROS)}, 2016, pp.
  751--758.

\bibitem{crosby2017integrating}
M.~Crosby, R.~P. Petrick, F.~Rovida, and V.~Krueger, ``Integrating mission and
  task planning in an industrial robotics framework,'' in \emph{Twenty-Seventh
  International Conference on Automated Planning and Scheduling}, 2017.

\bibitem{rovida2015design}
F.~Rovida and V.~Kr{\"u}ger, ``Design and development of a software
  architecture for autonomous mobile manipulators in industrial environments,''
  in \emph{IEEE International Conference on Industrial Technology (ICIT)},
  2015, pp. 3288--3295.

\bibitem{wantia2016task}
N.~Wantia, M.~Esen, A.~Hengstebeck, F.~Heinze, J.~Rossmann, J.~Deuse, and
  B.~Kuhlenkoetter, ``Task planning for human robot interactive processes,'' in
  \emph{IEEE 21st International Conference on Emerging Technologies and Factory
  Automation (ETFA)}, 2016, pp. 1--8.

\bibitem{stenmark2015connecting}
M.~Stenmark and J.~Malec, ``Connecting natural language to task demonstrations
  and low-level control of industrial robots.'' in \emph{MuSRobS@ IROS}, 2015,
  pp. 25--29.

\bibitem{scherzinger2019contact}
S.~Scherzinger, A.~Roennau, and R.~Dillmann, ``Contact skill imitation learning
  for robot-independent assembly programming,'' in \emph{IEEE/RSJ International
  Conference on Intelligent Robots and Systems (IROS)}, 2019, pp. 4309--4316.

\bibitem{kharidege2017practical}
A.~Kharidege, D.~T. Ting, and Z.~Yajun, ``A practical approach for automated
  polishing system of free-form surface path generation based on industrial arm
  robot,'' \emph{The International Journal of Advanced Manufacturing
  Technology}, vol.~93, no.~9, pp. 3921--3934, 2017.

\bibitem{schou2018skill}
C.~Schou, R.~S. Andersen, D.~Chrysostomou, S.~B{\o}gh, and O.~Madsen,
  ``Skill-based instruction of collaborative robots in industrial settings,''
  \emph{Robotics and Computer-Integrated Manufacturing}, vol.~53, pp. 72--80,
  2018.

\bibitem{cho2020learning}
N.~J. Cho, S.~H. Lee, J.~B. Kim, and I.~H. Suh, ``Learning, improving, and
  generalizing motor skills for the peg-in-hole tasks based on imitation
  learning and self-learning,'' \emph{Applied Sciences}, vol.~10, no.~8, p.
  2719, 2020.

\bibitem{schleipen2014automationml}
M.~Schleipen, J.~Pfrommer, K.~Aleksandrov, D.~Stogl, S.~Escaida, J.~Beyerer,
  and B.~Hein, ``Automationml to describe skills of production plants based on
  the ppr concept,'' in \emph{3rd AutomationML user conference}, 2014.

\bibitem{akkaladevi2019skill}
S.~C. Akkaladevi, A.~Pichler, M.~Plasch, M.~Ikeda, and M.~Hofmann,
  ``Skill-based programming of complex robotic assembly tasks for industrial
  application,'' \emph{e \& i Elektrotechnik und Informationstechnik}, vol.
  136, no.~7, pp. 326--333, 2019.

\bibitem{schlenoff2012ieee}
C.~Schlenoff, E.~Prestes, R.~Madhavan, P.~Goncalves, H.~Li, S.~Balakirsky,
  T.~Kramer, and E.~Miguelanez, ``An ieee standard ontology for robotics and
  automation,'' in \emph{2012 IEEE/RSJ international conference on intelligent
  robots and systems}.\hskip 1em plus 0.5em minus 0.4em\relax IEEE, 2012, pp.
  1337--1342.

\bibitem{de2019prototyping}
A.~B. de~Oliveira~Neto, J.~A. Silva, and M.~E. Barreto, ``Prototyping and
  validating the cora ontology: Case study on a simulated reconnaissance
  mission,'' in \emph{2019 Latin American Robotics Symposium (LARS), 2019
  Brazilian Symposium on Robotics (SBR) and 2019 Workshop on Robotics in
  Education (WRE)}.\hskip 1em plus 0.5em minus 0.4em\relax IEEE, 2019, pp.
  341--345.

\bibitem{verhoosel2017recipe}
J.~P.~C. Verhoosel and M.~A. {van Bekkum}, ``{Recipe-Based Engineering and
  Operator Support for Flexible Configuration of High-Mix Assembly},'' in
  \emph{{Advances in Production Management Systems. The Path to Intelligent,
  Collaborative and Sustainable Manufacturing}}, H.~L{\"o}dding, R.~Riedel,
  K.-D. Thoben, G.~von Cieminski, and D.~Kiritsis, Eds.\hskip 1em plus 0.5em
  minus 0.4em\relax Cham: {Springer International Publishing}, 2017, pp.
  363--371.

\bibitem{johansen2021role}
K.~Johansen, S.~Rao, and M.~Ashourpour, ``{The Role of Automation in
  Complexities of High-Mix in Low-Volume Production -- A Literature Review},''
  \emph{{Procedia CIRP}}, vol. 104, pp. 1452--1457, 2021.

\end{thebibliography}

\end{document}